\documentclass[conference]{IEEEtran}
\ifCLASSINFOpdf
\else
\fi

\usepackage{graphicx}
\usepackage{tikz}
\usepackage{amssymb}
\usepackage{amsmath}
\usepackage{todonotes}
\usepackage[normalem]{ulem}

\begin{document}
\title{ROS2-Based Simulation Framework for Cyberphysical Security Analysis of UAVs}

\author{
    \IEEEauthorblockN{Unmesh Patil, Akshith Gunasekaran, Rakesh Bobba, Houssam Abbas}
    \IEEEauthorblockA{\texttt{\{patilun, gunaseka, rakesh.bobba, houssam.abbas\}@oregonstate.edu}}
    \IEEEauthorblockA{Oregon State University}
}

\IEEEoverridecommandlockouts
\makeatletter\def\@IEEEpubidpullup{6.5\baselineskip}\makeatother

\maketitle
\newcommand{\PP}[1]{
    \vspace{1px}
    \noindent{\bf \IfEndWith{#1}{.}{#1}{#1.}}
}

\newcommand{\PPC}[2]{
    \vspace{5px}
    \noindent{{\bf #1} (#2):}
}

\newcommand{\PS}[1]{
    \vspace{5px}
    \noindent{\bf #1}
}

\newcommand{\PC}[2]{
    \vspace{5px}
    \noindent{\bf\CC{#1} \IfEndWith{#2}{.}{#2}{#2.}}
}

\newcommand*\CC[1]{%
    \begin{tikzpicture}[baseline=(C.base)]
        \node[draw,circle,inner sep=0.2pt](C) {#1};
    \end{tikzpicture}}

\newcommand{\one}{\CC{1}}
\newcommand{\two}{\CC{2}}
\newcommand{\three}{\CC{3}}

\begin{abstract}
    We present a new simulator of Uncrewed Aerial Vehicles (UAVs) that is
    tailored to the needs of testing cyber-physical security attacks and
    defenses. Recent investigations into UAV safety have unveiled various attack
    surfaces and some defense mechanisms. However, due to escalating regulations
    imposed by aviation authorities on security research on real UAVs, and the
    substantial costs associated with hardware test-bed configurations, there
    arises a necessity for a simulator capable of substituting for hardware
    experiments, and/or narrowing down their scope to the strictly necessary.
    The study of different attack mechanisms requires specific features in a
    simulator. We propose a simulation framework based on ROS2, leveraging some
    of its key advantages, including modularity, replicability, customization,
    and the utilization of open-source tools such as Gazebo. Our framework has a
    built-in motion planner, controller, communication models and attack models.
    We share examples of research use cases that our framework can enable,
    demonstrating its utility.
\end{abstract}

\section{Introduction}
\label{sec:introduction}

Uncrewed Aerial Vehicles (UAVs), commonly known as drones, are revolutionizing
the civilian and commercial domains. Over the past decade, UAVs have found
applications in surveillance, asset delivery, photography, disaster relief,
mapping, and more. The expanding utility of UAVs has been augmented by advances
in edge-computing capabilities contributing to improved sensor suites and
automation. UAV swarms, formation control algorithms, and networking and
communication protocols have become a prominent focus of academic research and
industry, with consistent improvements.

Given this widening deployment of UAVs, their vulnerability to cyber-physical
attacks~\cite{fainekos23stealthySTL} and their potential to cause physical damage is a pressing concern.
Security attacks ranging from jamming that prevents UAVs from receiving remote
commands to active attacks that allow for the complete takeover of UAVs have
been reported~\cite{UAV_attack_types}. The security landscape is highly
evolving, and will only get more complex as deployments of multiple UAVs
(swarms) become prevalent. This necessitates tools that can accelerate research,
development and deployment of security mechanisms for UAVs.

\begin{figure}[ht]
    \centering
    \includegraphics[width=\linewidth]{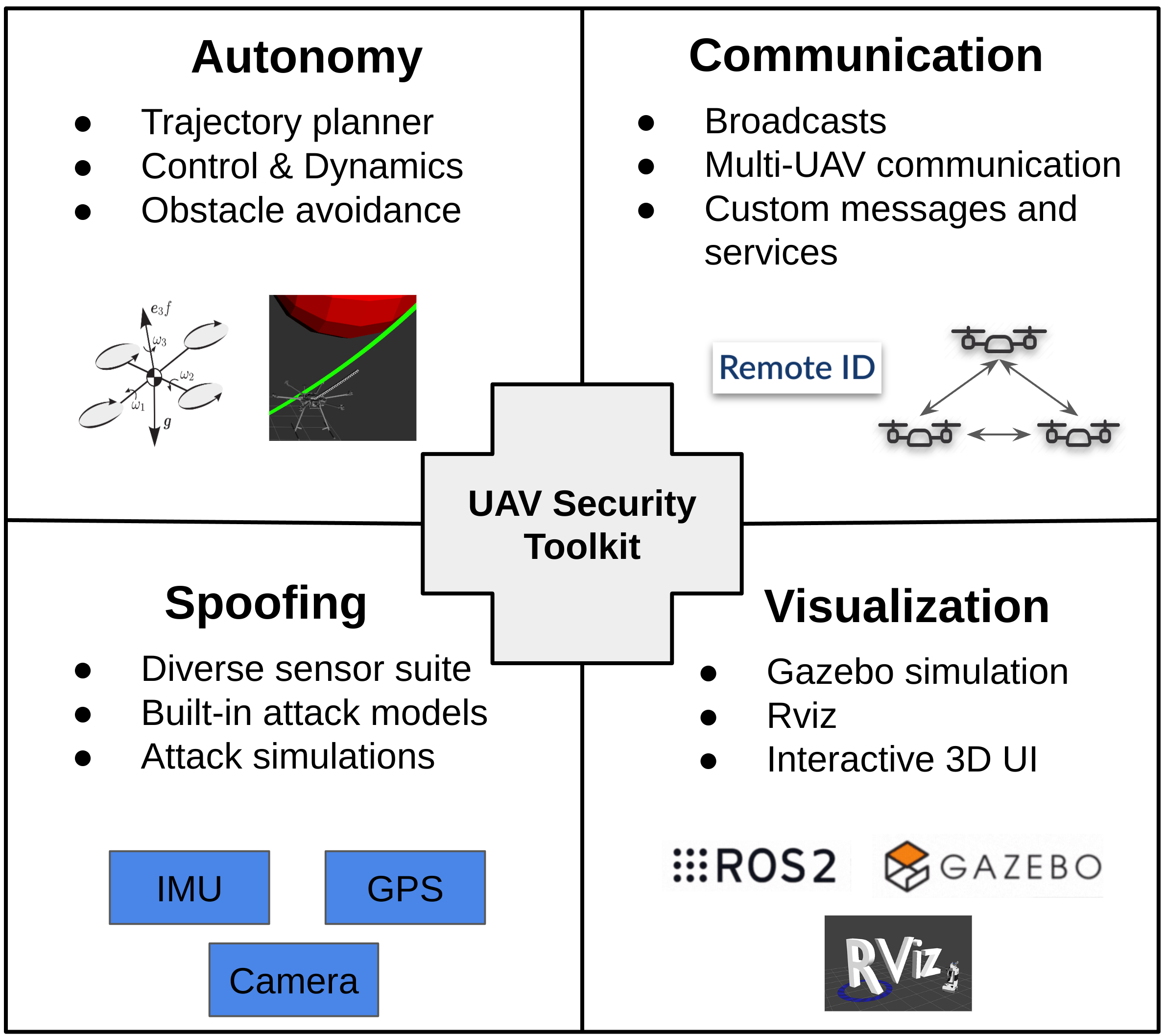}
    \caption{An overview of core features of the framework. Features are classified into four broad categories. All the listed features and corresponding examples are available off the shelf.}
    \label{fig:teaser}
\end{figure}
Governments worldwide have established regulatory authorities to govern the UAV
space like FAA (The US), EASA (The European Union), CAA (The UK). These
authorities are formulating stringent regulations governing UAV usage and
establishing standardized practices~\cite{regulation}. While these regulations
are essential, they impose significant costs on researchers studying
vulnerabilities for better future security. Furthermore, establishing a UAV
test-bed for single or multiple UAV testing is a financially demanding endeavor.
Consequently, researchers are relying on simulation-based studies more than
usual to evaluate the cyber-physical security of UAVs, narrowing the scope of
hardware experiments to strictly necessary.

The analysis of various types of attacks necessitates unique requirements from
simulation frameworks. The literature categorizes the attack surfaces of UAVs
into three broad classes~\cite{UAV_attack_types}: 1) Sensors, 2)
Computational/Control Units, and 3) Communications. Sensor attacks encompass
jamming or spoofing of GPS~\cite{gps_spoofing}, IMU~\cite{walnut, rockingdrones,
    unrocking}, Camera~\cite{laserspoofingDrew}, Altitude
sensors~\cite{altitude_sensor_KF_based}, among others. The second category
involves False Data Injection Attacks (FDIA)~\cite{false_data_injection} and
attacks on navigation or state-estimation
algorithms~\cite{altitude_sensor_KF_based}. Communication-related attacks
include Jamming, Spoofing, Eavesdropping, Interception, and Man-in-the-Middle
type attacks on various communication links. Furthermore, these attack surfaces
are also being actively studied in the context of UAV
swarms~\cite{swarmflawfinder},\cite{swarmfuzz}.

Within the literature, multiple endeavors by researchers are evident in creating
customized simulation frameworks tailored to their specific use cases. However,
this approach is time-consuming, adds costs to replicating the work, and
establishes a threshold requiring the learning a new simulator framework each
time. To address this issue, we propose an open-source, fully-customizable,
unified simulation framework based on the popular Robot Operating System 2
(ROS2)~\cite{ros2}. Figure~\ref{fig:teaser} shows major components and features
of the proposed framework. Our main contribution is the development of a ROS2
based simulation framework capable of simulating one or more autonomous UAVs and
a feature rich library of reusable components.
The code is available online at [address not shared in review copy for anonymity]
\textbf{Availability:} The framework is available for download at: https://github.com/patilunmesh/drone\_simulator

\section{Background and Related Works}
\label{sec:background}

Simulators have been employed to study various modules of Unmanned Aerial
Vehicles (UAVs) such as networks, sensors, state-estimation, swarm control
algorithms, and more. Popular simulators, such as WeBots \cite{webots_sim},
MORSE \cite{MORSE_sim}, CoppeliaSim (formerly Vrep) \cite{vrep_coppelia},
MRDS~\cite{mrds_microsoft}, ARGoS~\cite{argos}, USARSim~\cite{usarsim}, and
Gazebo+ROS~\cite{Gazebo_use}, among others, have been widely utilized. The task
of comparing and selecting the most suitable simulator for a specific use case
remains challenging. A comparison of some of the available simulators is shown
in Table 1. Notably, Farley et al.~\cite{how_to_pick} comprehensively compared
multiple simulators for mobile robots, elucidating their respective pros and
cons. Similarly, in~\cite{multiuav_sim_survey}, authors undertook a comparison
specific to multiple UAVs simulation, concluding that Gazebo + ROS emerges as
the most viable open-source option, considering factors such as physics engine
support, features, and customization capacity. We leverage this conclusion as
the foundation for the development of our simulation framework within the Gazebo
and ROS environment.

Studies in the literature aimed at analyzing the security of UAVs or multiple
UAVs have commonly involved the construction of custom simulation
frameworks~\cite{bunch_Equipment_sim, bunch_swarm_sim, bunch_swarm_testbed}. For
instance, in a study focused on effective countermeasures against UAV swarm
attacks~\cite{spoofing_sim_ros_gazebo}, authors engineered a custom framework
utilizing ROS and Gazebo. Javaid et al.~\cite{uavsim_security_sim} proposed a
simulation test-bed for UAV network cybersecurity analysis. Another example is
found in~\cite{bunch_RealtimeSS}, where a Matlab/Simulink-based simulation
system for UAVs is introduced. Souli et al.~\cite{multi-uav_ros_based_simulator}
developed a ROS-based communication network for simulating multiple UAVs and the
jamming of rogue UAVs.

However, these endeavors typically lack accessibility to their code-base (not
available for download), and even when provided, the custom nature of their
frameworks coupled with lacking reusability of the code introduces challenges to
replicability and future research endeavors. To address this issue, our
contribution involves the proposal of an open-source framework tailored for
security-related studies in the UAV domain. Our goal is to save efforts and time
spent on building simulation framework from scratch by establishing a reusable
code-base.

We underscore the importance of adopting the Robot Operating System 2 (ROS2) in
this context. ROS2 presents itself as an advanced middle-ware that builds upon
the strengths of its predecessor, ROS1. In the realm of UAV research, where
diverse modules and intricate interactions necessitate a robust, adaptable and
scalable~\cite{ros2_scalable} framework, ROS2 emerges as a compelling
choice~\cite{maes2}. The modularity inherent in ROS2 facilitates the

specific requirements of various use cases. Moreover, the integration of ROS2
with Gazebo ensures a comprehensive simulation environment, combining realistic
physics engines with the flexibility of ROS2's communication structure. It also
allows the support for Ardupilot and PX4 platforms~\cite{ardupilot_gazebo}.
\begin{table*}[!t]
    \caption{Comparison of UAV Simulators}
    \centering
    \begin{tabular}{|c|c|c|c|c|c|}
        \hline
        \textbf{Simulator} & \textbf{ROS Support}                 & \textbf{Language Support} & \textbf{Physics engine} & \textbf{License} & \textbf{Limitations}
        \\
        \hline
        Gazebo             & Default                              & C++, Python
                           & Multiple                             & Open Source
                           & Poor computational performance
        \\
        \hline
        MORSE              & Default                              & Python
                           & Bullet                               & Open Source
                           & Not being updated
        \\
        \hline
        CoppeliaSim        & Plugin                               & C++, Python, Lua
                           & Multiple                             & Proprietary
                           & Poor scalability~\cite{coppelia_bad}
        \\
        \hline
        Webots             & Plugin                               & C, C++, Python
                           & ODE                                  & Open Source
                           & Poor model format support
        \\
        \hline
        ARGoS              & Plugin                               & C++, Lua
                           & Multiple                             & Open Source
                           & Poor cross platform support
        \\
        \hline
    \end{tabular}
\end{table*}

\section{Design Goals}
\label{sec:design_goals}

\PS{Open and Modular Framework.} \emph{Open licensing} fosters transparency and
broad scrutiny while accelerating vulnerability discovery and patching, making
it ideal for rapid security research and development. In addition, it enables
\emph{easy replication} of existing research. While a fraction of existing
papers do publish their work with an open license, the lack of \emph{modular and
    reusable components} make it a challenge to extend their work. This motivates
our first design goal.
\PS{Robust Framework.} Finding a successful attack or defense involves running
numerous experiments over a period of time to identify vulnerable scenarios.
This requires the framework to have \emph{high computational performance} and
\emph{scale up} with more drones being added to the system. Additionally, giving
the developer a \emph{choice of language} will allow them to prioritize
performance or development time, whichever is suitable.

\PS{World Toggle.} Some tasks can be simulated without rendering the 3D scene by
assuming ideal world. Giving the research access to a low latency ideal world
will accelerate testing correctness of various control algorithms without the
overhead of rendering computationally expensive real world. This motivates an
optional rendering or offline mode in the framework.

\PS{Reusable library of components.} Simulation of autonomous UAVs involves
setting up components such as \emph{Trajectory Planner}, \emph{Controller} and
precise \emph{Dynamics model}. Additionally, \emph{sensor attack models} and
\emph{prediction models} are essential for security research. Making a library
of these features available off-the-shelf removes the burden of setup from the
user and improves the usability of the framework, letting them focus on specific
components. Other helpful utilities include support for \emph{Diverse sensor
    suite}, \emph{multi-UAV communication} and \emph{broadcasts}.

Within the proposed framework, we achieve \textbf{Openness and Modularity} by
using appropriate licensing and software engineering principles during the
development of the framework. The foundational elements of ROS2 and Gazebo help
us achieve high computational performance, scalability making out framework
\textbf{Robust}, in addition to enabling replicability via straightforward
recording and transfer of ROS2 bags.

Our main contributions center on addressing the last two goals. \one{} We
introduce an offline mode explained in section \ref{sec:framework-toggle} to
achieve World Toggle. \two{} We enhance the framework with an extensive library
of sensor attack models derived from security literature. The autonomy module,
offers off-the-shelf autonomous UAV simulation with a trajectory planner, LQR
controller and obstacle avoidance. Additionally, we provide examples for
multi-UAV communication setup and a Remote ID broadcast setup. We elaborate this
in section \ref{sec:framework-modules}.

\begin{figure}[ht]
    \centering
    \includegraphics[width=\linewidth]{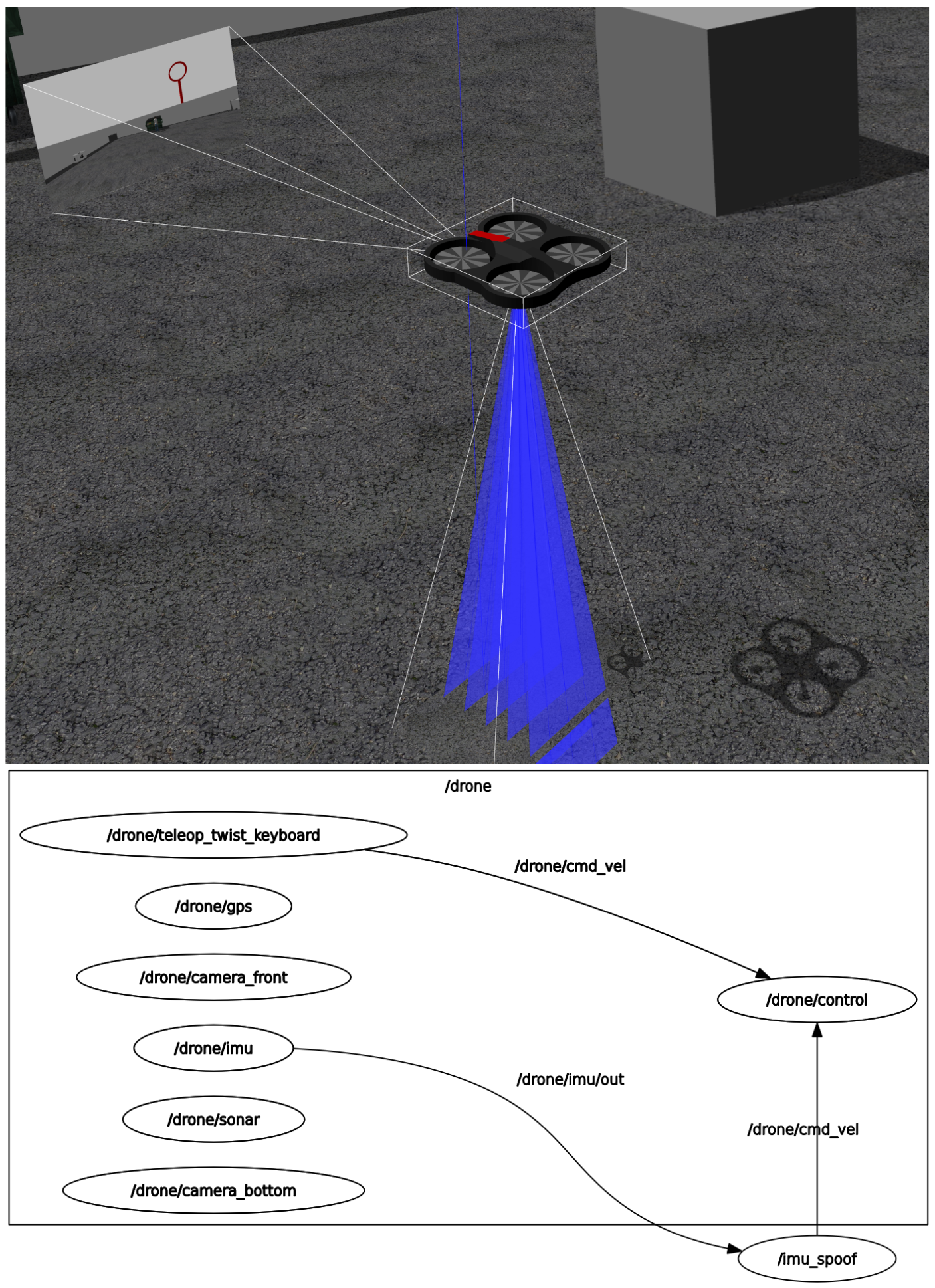}
    \caption{Online mode: The figure on the top shows a manually controlled UAV in Gazebo world which simulates two camera sensors (Front and bottom (blue colour)), IMU, GPS, and SONAR sensor. Figure in the bottom shows an example ROS2 Node graph for a simple IMU spoofing scenario. All the sensor topics are listed on the left. The imu\_spoof node subscribes to the IMU topic and calculates the effect of attack using attack model. This node also sends velocity commands to simulate the effect in real time.}
    \label{fig:online}
\end{figure}

\begin{figure}[ht]
    \centering
    \includegraphics[width=\linewidth]{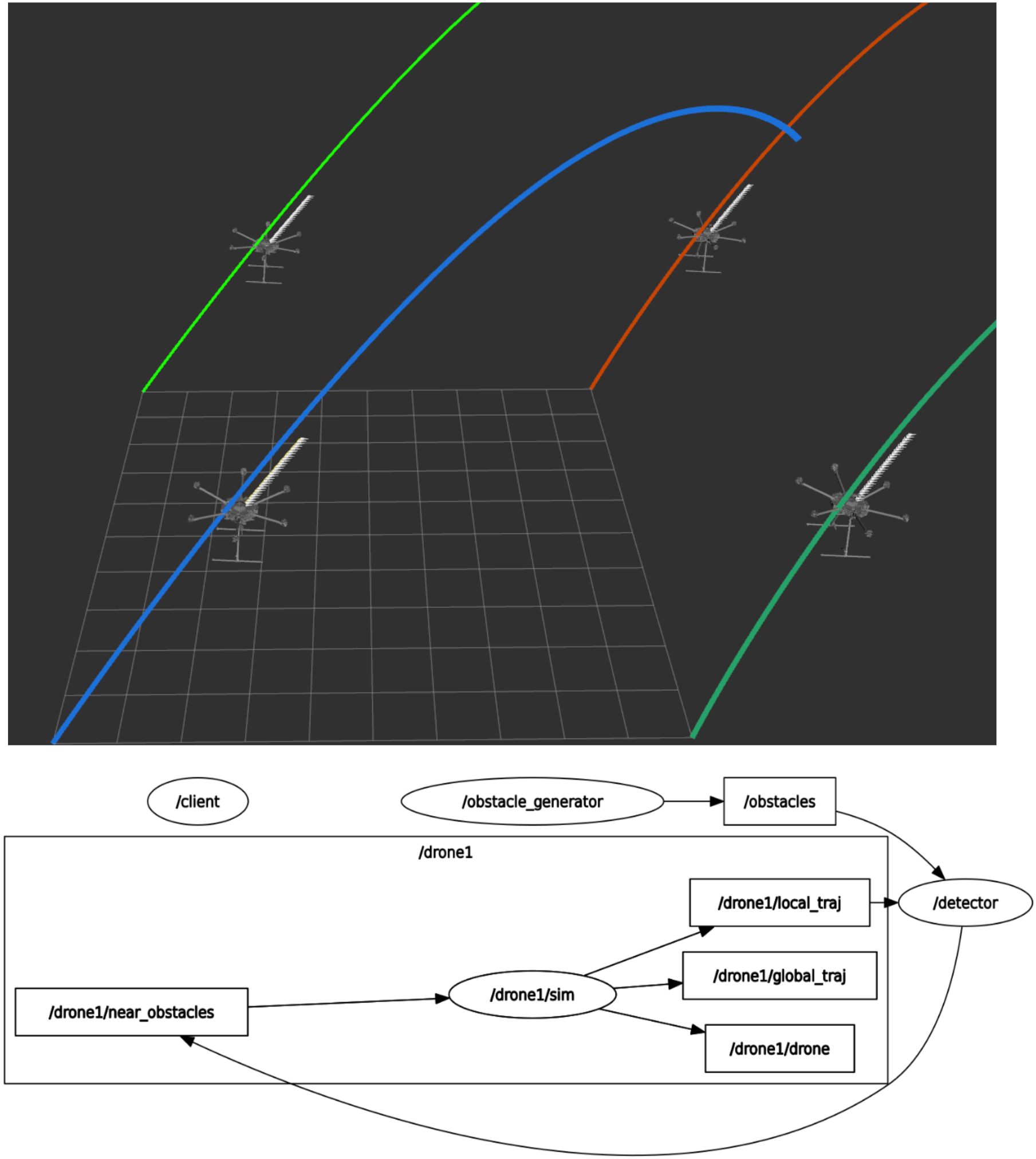}
    \caption{Offline mode: The figure on the top shows RViz visualization of four autonomous UAVs with global and local trajectories. The figure in the bottom is a ROS2 Node graph showing a service-client implementation of simulation. The drone1 sim service is called by a client node. The simulation node publishes global and local trajectories and subscribes to nearby obstacles. The obstacle generator node publishes obstacles, which are then detected by a detector node. This bloack repeats for each UAV.}
    \label{fig:offline}
\end{figure}

\section{Simulation Framework}
\label{sec:framework}

\subsection{World Toggles}
\label{sec:framework-toggle}

\PS{Online Mode.} In this mode, UAVs are simulated within Gazebo environment to
generate realistic sensor data as shown in Figure~\ref{fig:online}. There are multiple implementations of UAV models
for Gazebo available online. These models can be easily plugged in with the
current framework. This is primarily designed to study in real-time, sensor
spoofing attacks. We provide a library of various sensor spoofing attacks as
ROS2 nodes. We also provide various attack-models that allow researchers to
customize the effects of sensor spoofing.

\PS{Offline Mode}: This mode is tailored for testing algorithms under ideal
conditions. We compute the trajectory and control inputs beforehand and simulate
an algorithm which is then visualized by Rviz as shown in Figure~\ref{fig:offline}. Devoid of a rendering or physics
engine tool like Gazebo, this mode employs non-linear UAV
dynamics~\cite{sabatinothesis}, an LQR controller, and a rapid trajectory
planner~\cite{rapid_trajectory}, implemented as Python scripts running as ROS2
Nodes. These modules can also be utilized in Online mode. It is particularly
useful for testing obstacle avoidance algorithms, swarm control algorithms, and
trajectory planners.

\subsection{Modules}
\label{sec:framework-modules}

The simulation framework incorporates several modules, each contributing
distinct functionalities. A brief summary of available features is shown in
Table~\ref{tab:features}.  This section provides an exhaustive overview of each
module, accompanied by illustrative use cases.\\

\begin{table}[ht]
    \centering
    \caption{Available modules and features of the framework}
    \begin{tabular}{|c|c|}
        \hline
        Module        & Available features \\
        \hline
        Sensors       & IMU, GPS, Cameras, SONAR \\
        \hline
        Autonomy      & Trajectory planner, LQR controller, Obstacle avoidance
        \\
        \hline
        Attack models & IMU, GPS, Camera \\
        \hline
        Communication & Multi-UAV communication, Remote ID broadcast \\
        \hline
        Others        & Obstacle publisher and detector, prediction models \\
        \hline
    \end{tabular}

    \label{tab:features}
\end{table}

\textbf{Trajectory Planner:} The Trajectory Planner module optimizes for a given
objective, generating a set of states or a plan based on the initial and goal
poses, as well as motion constraints. State of the UAV can be represented by 12
dimensional vector as, $[x, y, z, \theta, \psi, \omega, \dot x, \dot y, \dot z,
\dot \theta, \dot \psi, \dot \omega]^T$. These fields represent position,
orientation (euler angles), linear velocities and angular velocities along 3
axes. Input parameters include the 12-dimensional vectors for start and goal
poses, the total mission time, and temporal resolution. More details on motion
constraints and planning objectives can be found in~\cite{rapid_trajectory}. The
output, presented as lists of state variables at a specified temporal
resolution, is converted into a ROS2 topic with nav-msgs Path type message. We
provide a trajectory planner with the framework but multiple such planners are
available as ROS2 packages and can be utilized in the same fashion.

\textbf{Controller:} In the current implementation, LQR controller is provided
to generate control inputs based on the trajectory generated by the planner
module. The controller module utilizes non-linear dynamics and ODEINT solver
from Scipy library. The set of dynamics equations in the current setting are
listed below, for detailed derivation of these dynamics equations please refer
to~\cite{sabatinothesis}.

\textbf{Dynamics equations:}
\begin{center}
    $\begin{array}{c}

            {\ddot{x}=}{-\frac{f_{t}}{m}[s(\phi)s(\psi)+c(\phi)c(\psi)s(\theta)]}
            \\\\

            {\ddot{y}=}{-\frac{f_{t}}{m}[c(\phi)s(\psi)s(\theta)-c(\psi)s(\phi)]}
            \\\\

            {\ddot{z}=}{g - \frac{f_{t}}{m}[c(\phi)c(\theta)]} \\\\

            {\ddot{\phi}=}{\frac{I_{y}-I_{z}}{I_{x}}\dot{\theta}\dot{\psi}+\frac{\tau_{x}}{I_{x}}}
            \\\\

            {\ddot{\theta}=}{\frac{I_{z}-I_{x}}{I_{y}}\dot{\phi}\dot{\psi}+\frac{\tau_{y}}{I_{y}}}
            \\\\

            {\ddot{\psi}=}{\frac{I_{x}-I_{y}}{I_{z}}\dot{\phi}\dot{\theta}+\frac{\tau_{z}}{I_{z}}}
        \end{array}$ \\
\end{center}
Where, $\phi, \theta, \psi$ are euler angles and $c(\phi) = cos(\phi)$. $f_t$ is
the thrust force and $m$ is mass of the UAV. $I_x, \tau_x$ represent inertia and
control torques about the corresponding axes respectively. The result of control
output is then visualized in Rviz for offline mode as a local plan. The control
output or the planned trajectory can be used for Online mode with Gazebo as
well.

\textbf{Obstacle manager:} This module generates dynamic and static obstacles.
The current code shows a demo of publishing static and dynamic obstacles as Rviz
markers. Obstacle detectors are also available in the framework with an option
to choose global Vs local detection. Local detection lets you select the radius
within which the obstacles are considered as nearby and once the obstacle is
detected it is published on nearby-obstacles topic. Obstacle avoidance is
implemented in a repulsion-like fashion, and has a room for modification by
implementing various algorithms.

\textbf{Prediction models:} For tracking and predicting the future path of an
unknown UAV, the framework offers multiple prediction models. Ranging from a
linear interpolation model with a first-order Markov assumption to
dynamics-aware and obstacle-aware models, the suite provides flexibility to
researchers. Additionally, an Extended Kalman Filter template for prediction is
provided.

\textbf{Broadcasts \& communications:} Critical for Multi-UAV simulation, this
module enables communication between multiple UAVs, allowing researchers to
customize interactions and introduce broadcasts. In the current simulation
framework, we provide custom messages to simulate such broadcasts. A Remote ID
broadcaster module is included, along with a custom message type designed as per
the standard protocol that serves as a guide for creating custom communication
modules.

\textbf{Attack models:} Attack models define how the UAV system reacts to sensor
spoofing. Different attack surfaces have different effects on the UAV system.
For example, an acoustic injection attack~\cite{walnut} on accelerometer sensor
can induce a change in the state estimation of the UAV, however the amount of
change it can induce is hugely limited by onboard recursive filters. In addition
to that, the attack mechanism allows only intermittent access to the attack
surface. All these observations are captured in an attack model of IMU sensor to
simulate the effect of attack on the sensor realistically. Our framework has a
range of built-in attack models for IMU, GPS and Camera sensors. These models
are available on download and can be easily modified as per the need.

\subsection{Implementation:}
The modules explained in the preceding section represent nodes in the framework.
Nodes are fundamental process units within ROS2, Nodes execute discrete
functionalities, contributing to the modular structure of the framework. Nodes
can employ various interfaces. These interfaces encompass Topics, Services, and
Actions. Another feature for implementation is ROS2 plugins, which is seamlessly
integrated via \texttt{pluginlib}. Each interface serves distinct target
applications:

\textbf{Topics:} Employed for continuous data streams, Topics facilitate the
exchange of real-time information between nodes. Topics convey messages in
specific formats. Messages are data structures with both predefined and
customizable data types. All sensor data streams are implemented in this way.

\textbf{Services:} Grounded in a server-client protocol, Services are used for
remote process calls towards short-lived interactions. The simulation node is
implemented using this interface. For sequential decision-making approaches, a
service client model of simulation allows for a request-response type of
setting.

\textbf{Actions:} Suited for invoking prolonged processes, Actions provide the
option for continuous feedback and yield results upon completion. Actions are
basically a combination of topic and services. This mode is not utilized in the
current code but can be easily accommodated.

\begin{figure}[ht]
    \centering
    \includegraphics[width=0.7\linewidth]{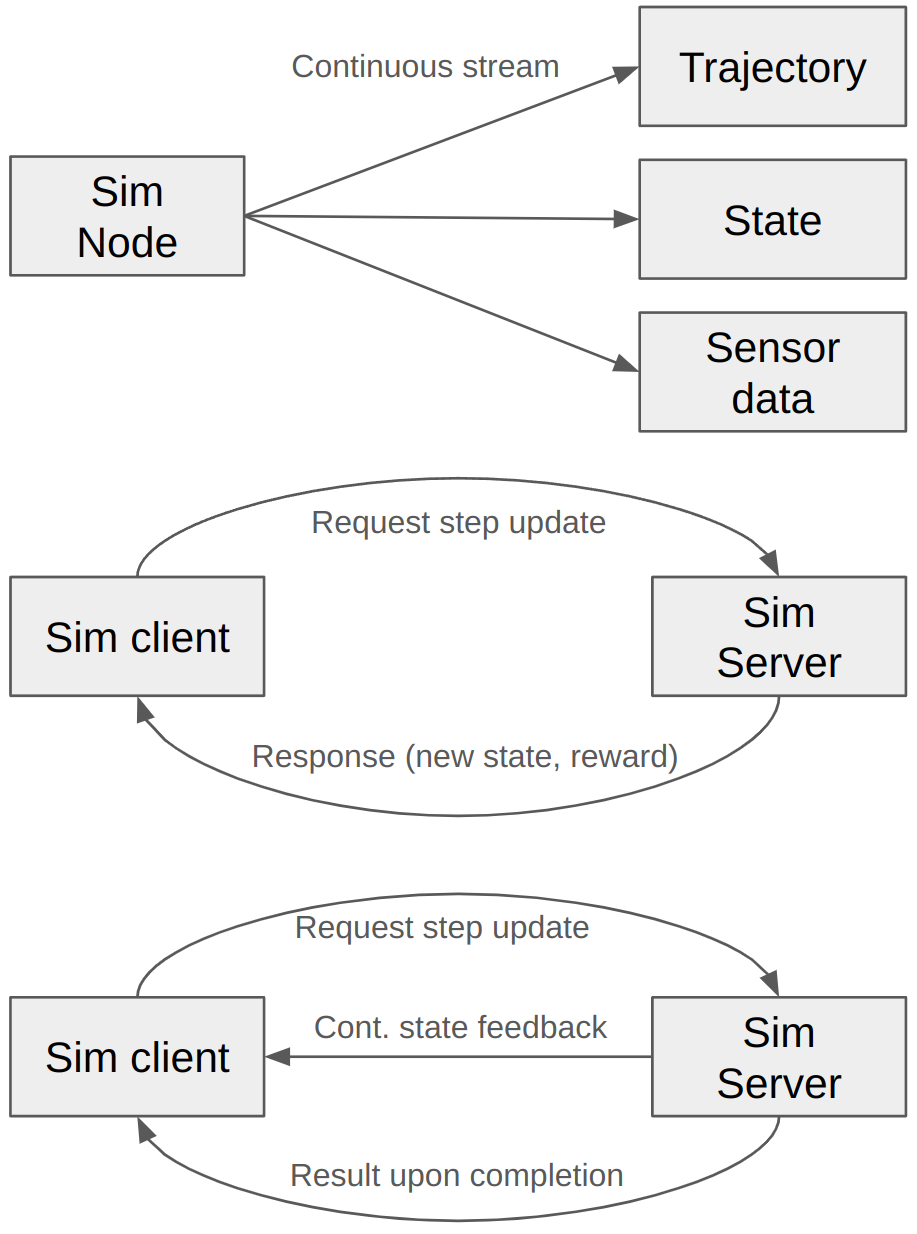}
    \caption{Overview of different interfaces used in the simulation framework. The flowchart on the top shows the topic like structure of the framework, the second flowchart indicates a simple server-client structure for the simulation. The last flowchart shows the typical structure of a ROS2 action.}
    \label{fig:implement}
\end{figure}

The core simulator node is implemented as both a service and a topic.
Figure~\ref{fig:implement} shows the working mechanism of each interface.

\subsection{Extensions}

ROS2 is known for their community support and off-the-shelf packages. Most of
these open source modules can potentially be used alongside our simulation
framework. A few examples include: visual SLAM package used for attacks on
visual navigation can easily be added to the existing platform. Similarly, QR
code based navigation or QR-code landing can be simulated in Gazebo using ROS2
packages. Swarm control algorithms can also be added to extend the library.

\section{Use Cases}
\label{sec:usecases}

In this section we present a few use cases of the UAV Security Toolkit.

\subsection*{GNSS Signal Jamming}

One of the biggest threats to drone safety is GNSS interference. In the best
case scenario, disruption of satellite signal can degrade the position quality
leading the drone to fall back from high precision to low precision positioning
using other sensors. In the worst case, interference can cause complete loss in
tracking and positioning. Defenses have been developed to detect and correct for
such scenarios~\cite{gps_jamming_survey}. An essential tool required in
developing such defenses is a Jamming Simulator which essentially mimics
interference in the GPS signal.

The framework supports jamming using ROS2's QoS and Message publishing rate.
Multiple Jamming strategies such as 1. User Controlled 2. Location
Based 3. Time based or 4. Combination allows for testing complex jamming
scenarios and corresponding defense strategies without having to build a jamming
simulator from scratch or having to resort to expensive hardware jammers. In
addition, the jammer is agnostic to signal/communication type.

\subsection*{Spoofing Multiple Drones}

Spoofing is a growing threat where the attack sends fake signals to one or more
drones to manipulate their navigation and cause them to crash. It can involve
sending fake GNSS signals or creating fake obstacles in the drone's path. The
framework supports spoofing by allowing the user to specify the signal to be
spoofed along with parameters such as time, location and duration of the spoof.
Additionally, the framework allows for setting a threshold for the spoofed
signal, this will limit the spoofing to a certain range to avoid detection. With
respect to obstacle, it allows for custom obstacles to be created or imported.
It also allows for spoofing multiple drones at the same time, which is necessary
for testing swarms where the drones are in proximity to each other and are
susceptible to the same spoofing attack.

This enables testing robustness and correctness of the drone's navigation system
and collision avoidance system against spoofing attacks and also test the
effectiveness of spoofing defenses.

\subsection*{Privacy analysis of Remote ID}

This use case demonstrates the ease of implementing new message formats in the
framework. The Remote ID was recently proposed by the FAA as a means to identify
drones in the airspace. Works such as ~\cite{remote_id_privacy} have
investigated privacy issues with Remote ID such as the ability to track a
drone's flight path and the ability to identify the drone's owner. The framework
allows for implementing Remote ID message by simply defining the message format
as a ROS2 message and implementing the message publisher, without having to wait
for any reference implementation to be released or hardware to be available.

\section{Conclusion}
\label{sec:conclusion}
In conclusion, we introduce a simulation framework for Unmanned Aerial Vehicles
(UAVs) with a dedicated emphasis on security research. Leveraging the
versatility of the Robot Operating System 2 (ROS2), our framework encompasses
critical functionalities, including a Trajectory Planner, Controller, Obstacle
Manager, Prediction Models, Broadcasts \& Communications, and Attack Models.
These modules collectively form a comprehensive platform that addresses current
challenges in UAV security research and anticipated future developments.
Implementation is facilitated through ROS2 interfaces, emphasizing modularity
and adaptability. Our open-source framework provides a standardized environment
for UAV security analysis and invites collaborative contributions to further
refine and enhance its capabilities.

Looking ahead, we foresee continuous evolution and refinement of the framework,
fostering innovation in UAV security research. With this platform, we aim to
catalyze advancements in understanding security scenarios, assessing attack
impacts, and developing resilient countermeasures for UAV systems.

\bibliographystyle{IEEEtranS}

\bibliography{refs}

\end{document}